\documentclass[preprint,12pt]{elsarticle}

\usepackage{amssymb}
\usepackage{mathtools}
\usepackage{url}
\usepackage{amsmath,amssymb,amsfonts}
\usepackage{algorithmic}
\usepackage{graphicx}
\usepackage{textcomp}
\usepackage{xcolor}
\usepackage{fancyhdr}
\usepackage{xfrac}
\usepackage[justification=centering]{caption}
\usepackage{float}
\restylefloat{table}
\usepackage{caption}
\usepackage{subcaption}

\usepackage[final]{changes}

\newcommand\Mycomb[2][^n]{\prescript{#1\mkern-0.5mu}{}C_{#2}}

\journal{Microprocessors and Microsystems}

\makeatletter
\def\ps@pprintTitle{%
 \let\@oddhead\@empty
 \let\@evenhead\@empty
 \def\@oddfoot{}%
 \let\@evenfoot\@oddfoot}
\makeatother

\begin{document}

\begin{frontmatter}



\title{On Extending Amdahl's law to Learn Computer Performance}


\author[inst1]{Chaitanya Poolla}
\author[inst1]{Rahul Saxena}

\affiliation[inst1]{organization={Intel Corporation},
            addressline={3600 Juliette Ln}, 
            city={Santa Clara},
            state={CA 95054},
            country={USA}}



\begin{abstract}
The problem of learning parallel computer performance is investigated in the context of multicore processors. Given a fixed workload, the effect of varying system configuration on performance is sought. Conventionally, the performance speedup due to a single resource enhancement is formulated using Amdahl's law. However, in case of multiple configurable resources the conventional formulation results in several disconnected speedup equations that cannot be combined together to determine the overall speedup. To solve this problem, we propose to (1) extend Amdahl's law to accommodate multiple configurable resources into the overall speedup equation, and (2) transform the speedup equation into a multivariable regression problem suitable for machine learning. Using experimental data from \added{fifty-eight tests spanning} two benchmarks (\emph{SPECCPU 2017 and PCMark 10}) and four hardware platforms (\emph{Intel Xeon 8180M, AMD EPYC 7702P, Intel CoffeeLake 8700K, and AMD Ryzen 3900X}), analytical models are developed and cross-validated. Findings indicate that in most cases, the models result in an average cross-validated accuracy higher than $95\%$, thereby validating the proposed extension of Amdahl's law. The proposed methodology enables rapid generation of \deleted{intelligent} \added{multivariable} analytical models to support future industrial development, optimization, and simulation needs.
\end{abstract}



\begin{keyword}
Computer performance \sep analytical modeling \sep Amdahl's law \sep machine learning \sep regression
\end{keyword}

\end{frontmatter}



\section{Introduction}
\label{sec:introduction}
Modern computers consist of complex interdependent subsystems working in tandem to provide a seamless user experience. These subsystems realize a variety of underlying implementations and architectures in hardware \cite{gonzalez2010processor}. Computer performance is evaluated by benchmarking, a process that subjects computer systems to standard load conditions and monitors performance metrics of interest \cite{leboudec2010performance}. Depending on the end goal, performance metrics could represent aspects such as response time, throughput, or power consumption.

The system performance can be obtained by different methods such as collecting measurements, running simulations, or predicting from analytical models \cite{jain2008art}. In case of measurements, the workload is configured to run on the test system under standard conditions and the performance is reported. While the accuracy could be relatively high, the experimenter has limited feasible configurations of the system to choose from. In case of simulations, the \deleted{subsytems}\added{subsystems} and their interactions can be modeled extensively. Depending on the complexity of the simulator, there is a trade off between accuracy, coverage, development time, and evaluation time. For example, complex simulators often require more time to achieve a higher accuracy, and simpler simulators require less time at the expense of lower accuracy.
Further, simulators require extensive validation for reliable use which is time-consuming \cite{eeckhout2010computer} \cite{akram2019survey}. However, simulators allow for investigating performance of next generation systems with hypothetical system configurations. 

Given the limited scope of measurement configurations and the complex nature of simulations, it is of interest to explore simpler analytical models partly based on benchmark measurements. Such models allow one to investigate a larger scope of system configurations in a short time and hence blend the advantages of both measurement and simulation-based methods. Previous research suggests that easy to use and understandable analytical models that can explore large design spaces are imperative for architectural analysis \cite{skadron2003challenges} \cite{van2015micro}. In this study, we focus on the problem of developing analytical models for parallel computing in the context of multi-core processors based on limited benchmark measurements.

For a given benchmark, the performance of a multi-core processor depends on several factors related to the system configuration, hardware architecture, and the software stack used \cite{hoefler2015scientific}. After fixing the hardware and software stack, measurements are obtained by running the benchmark at a specified system configuration. The system configuration is varied by variables such as the number of cores, threads per core, core frequency, memory frequency, and cache sizes \cite{ipek2006efficiently}. Each of these variables is set to a feasible value prior to the execution of the benchmark. Thus, each system configuration can be viewed as an instance of the corresponding machine. Accordingly, the same computer hardware can be used to build several system configurations each of which provides a corresponding set of system resources for executing the benchmark.

The execution of a benchmark on a given system configuration generates performance values, commonly known as scores. These scores allow for comparison of performance across different configurations, architectures, and software stacks for a common benchmark. Typically, a baseline and a test system are chosen for executing the benchmark to obtain the baseline and test score respectively. When the benchmark performance metric measures the execution time or latency, the ratio of the baseline to test scores provides the performance speedup on the test system in comparison to the baseline system.

The theoretical speedup expected from a test system with improved resources is provided by Amdahl's law \cite{amdahl1967validity} \cite{hill2008amdahl}. In case of multi-core processors, the speedup could result from scaling up a system resource such as the number of cores. In the conventional formulation of the Amdahl's law, a program is comprised of two parts: A parallelizable part whose performance is affected by scaling up system resources and a sequential part whose performance remains unaffected by scale. \added{It is important to note that the problem size or workload remains fixed in the formulation of Amdahl's law. If the problem size increases parallely with the resource, Gustafon's model is to used to compute speedup \cite{gustafson1988reevaluating}. In this work, we assume a fixed problem size within the scope of Amdah's law.}. Over the recent years, variants of the Amdahl's law have been proposed to consider other aspects of system such as the memory wall problem \cite{sun2010reevaluating} or the synchronization problem \cite{eeckhout2010computer}. \added{More recently, a survey of methods to improve the semantic power of the p-fraction and computation capability improvement index can be found in \cite{al2020amdahl}}. It is important to note that while these variants are based on sound theoretical considerations they lack corroboration to empirical data from silicon measurements. Further, in these cases Amdahl's law is primarily employed to predict the theoretical speedup by changing one system resource at a time. This results in several equations, each corresponding to scaling an isolated system resource given fixed values of other system resources. Since each equation stems from different assumptions regarding the values of the other system resources, it is not feasible to combine these equations to account for the effect of multiple system variables simultaneously \cite{james2013introduction}. Thus, there is a need for a framework that can account for the effects of multiple variables simultaneously.

To this end, we propose an extension to Amdahl's law by adopting a multi-variable data-driven approach. From the viewpoint of program execution, the need for this extension arises because real-world applications view the system not as comprised of isolated resources but as multiple resources interacting together to support the application. Therefore, the modeling procedure requires integrating these interaction effects. Further, such an extension allows data from real-world experiments to determine the model coefficients and hence offer insights on both the parallelizable fractions and expected scores for new configurations. In this manner, the extension accommodates multiple system resources simultaneously and offer a medium for experimental data to predict system performance on new system configurations. Related works on regression-based analytical modeling are found in \cite{lee2006accurate} and \cite{jia2012stargazer}. In \cite{lee2006accurate}, a regression-based predictive modeling approach is proposed consisting of less than 30 predictors, human-specified interactions, and 2000-4000 simulated observations. Each predictor was chosen based on the strength of its marginal correlations without accounting for the variation in other predictors. In the case of Stargazer \cite{jia2012stargazer}, a forward selection-based stepwise regression was employed to study the GPU design space tradeoffs. The study used 10 architectural parameters and 30-300 data points to achieve 85-99\% accuracy based on sample-train-test method. The use of stepwise regression is susceptible to model bias due to incorrect specification in the best subset selection problem. Further, both regression-based studies do not employ cross-validation, resulting in lack of unbiased generalization error. By contrast, we employ Amdahl's law-based models to achieve an average of $\approx95\%$ cross-validation accuracy with relatively fewer data points (20-200) given seven independent resources variables. To the best of the authors knowledge, this is the first work to reveal a direct connection between multivariable Amdahl's law and machine learning-based modeling. The contributions of this work are as follows:
\begin{enumerate}
    \item An extension of Amdahl's law to accommodate simultaneous scalings across multiple system resources and their interactions. 
    \item Statistical regression techniques to generate multi-variable analytical models based on measurements data.
    \item A demonstration of cross-validation and prediction efficacy of the proposed analytical models based on data from experiments with industry standard benchmarks and hardware.
\end{enumerate}
The remainder of the paper is structured as follows: Section \ref{sec:methodology} proposes the methodology based on Amdahl's law and its reformulation into a multiple regression framework. Section \ref{sec:experiment} describes the experiment setup including the Systems Under Test (SUTs) and the benchmarks considered. Section \ref{sec:modeling} discusses the model design, training, and validation. Section \ref{sec:results} presents the results along with the scope and modeling assumptions followed by the conclusion in Section \ref{sec:conclusion}.









\section{Methodology}
\label{sec:methodology}
Consider a system with configuration denoted by
$\mathcal{C}$. Let this configuration be specified by $k$ resource variables represented by the sequence $\left\{r_i^\mathcal{C}\right\}_{i=1}^k \subset {\mathbb{R}^+}^k$. Within a multi-core processor environment, these resource variables may represent core count, multi-threading state, core frequency, cache size, memory frequency, and other configurable variables. For a given hardware architecture, benchmark and software stack, let the baseline and the test system configurations be represented by $\mathcal{C}_b$ and $\mathcal{C}_t$, respectively. Accordingly, the baseline and test system resources are specified by $\left\{r_i^{\mathcal{C}_b}\right\}_{i=1}^k$ and $\left\{r_i^{\mathcal{C}_t}\right\}_{i=1}^k$, respectively. Further, let the baseline and test performance be specified by $P^{\mathcal{C}_b} \in \mathbb{R}^+$ and $P^{\mathcal{C}_t} \in \mathbb{R}^+$, respectively.
\subsection{Amdahl's Law: \deleted{Single resource enhancement}\added{Enhancement due to a single resource}}
\label{ssec:amdahl_single}
Given the above notation, let $f_i$ denote the fraction of the program enhanced exclusively by the $i^{th}$ system resource. Assuming the performance metric is the inverse of the execution time, we can employ Amdahl's law \cite{hill2008amdahl} the derive the theoretical performance speedup as:
\begin{equation}
\label{eqn:speedup_i}
    \mathcal{S}(i,f_i,\mathcal{C}_b,\mathcal{C}_t) = \frac{P^{\mathcal{C}_t}}{P^{\mathcal{C}_b}}\Bigg\rvert_{i^c} = \frac{1}{(1-f_i) + f_i\Bigg(\frac{r_i^{\mathcal{C}_b}}{r_i^{\mathcal{C}_t}}\Bigg)}
\end{equation}
where, $\mathcal{S}$ denotes the performance speedup, $i$ denotes the index of the $i^{th}$ resource, and $r_i^{\mathcal{C}_b}$, $r_i^{\mathcal{C}_t}$ denote the $i^{th}$ resource of the baseline and test systems, respectively. The symbol $\big\rvert_{i^c}$ refers to the operating condition that all resources other than the $i^{th}$ are equal across the baseline and test systems. The ratio $\sfrac{r_i^{\mathcal{C}_t}}{r_i^{\mathcal{C}_b}}$ denotes the resource enhancement. We note that in most cases even the parallel fraction $f_i$ is not infinitely parallelizable \cite{lotfi2018dark}. In other words, improving the $i^{th}$ resource by a factor of $x$ does not necessarily result in an equivalent speedup even within the parallelizable fraction $f_i$ of the program. Nevertheless, the speedup equation offered by Amdahl's law is useful to formulate the effect of parallelism stemming from resource enhancements. Further, we note that the above equation only considers the speedup resulting from an enhancement exclusively due to the $i^{th}$ resource - \emph{at fixed values of the other resources}. Similarly, the speedup resulting from enhancing only the $j^{th}$ resource may be given by:
\begin{equation}
\label{eqn:speedup_j}
    \mathcal{S}(j,f_j,\mathcal{C}_b,\mathcal{C}_t) = \frac{P^{\mathcal{C}_t}}{P^{\mathcal{C}_b}}\Bigg\rvert_{j^c} = \frac{1}{(1-f_j) + f_j\Bigg(\frac{r_j^{\mathcal{C}_b}}{r_j^{\mathcal{C}_t}}\Bigg)}
\end{equation}
\subsection{Amdahl's law: Need for reformulation}
\label{ssec:need_to_extend_amdahls_law}
Given the performance speedups exclusively due to $i^{th}$ and $j^{th}$ resources in equations \ref{eqn:speedup_i} and \ref{eqn:speedup_j} respectively, it is of interest to determine the overall speedup due to enhancing both $i^{th}$ and $j^{th}$ resources simultaneously. The overall speedup is particularly significant as it allows one to examine the effect of varying both resources simultaneously, unlike equations \ref{eqn:speedup_i} or \ref{eqn:speedup_j} which hold resources $j$ or $i$ at fixed values, respectively. From a statistical viewpoint, equations \ref{eqn:speedup_i} and \ref{eqn:speedup_j} can be viewed as simple linear regression models of the inverse of the performance speedup as a function of the $i^{th}$ and $j^{th}$ resource enhancements, respectively. The overall speedup equation due to varying resources $i$ and $j$ would need to combine the performance speedups due to the individual resource enhancements and the interactions between them. Unfortunately, there is no well-defined mechanism to combine the individual performance speedups in equations \ref{eqn:speedup_i} and \ref{eqn:speedup_j} directly to obtain the overall performance speedup \cite{james2013introduction}. Hence, we reconsider the Amdahl's law formulation to be able to incorporate multiple resource variables.
\subsection{Amdahl's Law: \deleted{Multiple resources enhancements}\added{Enhancement due to mutually exclusive resources}}
\label{ssec:amdahl_multiple}
Consider a scenario where mutually exclusive fractions of the program are enhanced by the speedups from the individual resources. In this case, the resources corresponding to the indices $1, 2, \cdots, k$ enhance the respective fractions $f_1, f_2, \cdots, f_k$ of the program. Accordingly, the performance speedup may be expressed as:
\begin{equation}
\label{eqn:speedup_m}
    \frac{P^{\mathcal{C}_t}}{P^{\mathcal{C}_b}} = \mathcal{S}(\cdot) = \frac{1}{(1-f_1-f_2-\cdots-f_k) + \sum\limits_{m=1}^k f_m\Bigg(\frac{r_m^{\mathcal{C}_b}}{r_m^{\mathcal{C}_t}}\Bigg)}
\end{equation}
where $\mathcal{S}(\cdot)$ denotes the shorthand notation of the multi-resource speedup function $\mathcal{S}\Big(\left\{i\right\}_{i=1}^k,\left\{f_i\right\}_{i=1}^k,\mathcal{C}_b,\mathcal{C}_t\Big)$.
\subsubsection{Enhancements due to resource interactions}
\label{ssec:amdahl_interactions}
The above formulation combines the individual effects of several resources into the overall performance speedup. However, during the program execution, interaction between multiple resources occur. Thus, the performance speedup formulation must include the effects of these resource interactions\footnote{In the context of experimental design, resource variables are known as factors. Thus, an interaction between $p\ (\leq k)$ resources is referred to as a \emph{p-factor interaction}}. Theoretically, interactions can occur between a minimum of two factors and a maximum of $k$-factors, thereby resulting in possibly $\Mycomb[k]{2} + \Mycomb[k]{3} + \cdots + \Mycomb[k]{k} = 2^k-k-1$ unique combinations, which is exponential in the number of individual resources $k$. In order to deal with this exponential complexity, we resort to the \emph{hierarchical ordering principle} whereby higher order interactions involving three or more factors are deemed very rarely significant and hence ignored \cite{hamada2000experiments}.
Accordingly, the speedup equation based on Amdahl's law with single and two-factor resource interactions can be expressed as:
\begin{multline}
\label{eqn:speedup_interactions}
    \frac{P^{\mathcal{C}_t}}{P^{\mathcal{C}_b}} = \mathcal{S}(\cdot) =  \frac{1}{\Bigg(1-\sum\limits_{m=1}^k f_m-\sum\limits_{m=2}^k\sum\limits_{n=1}^{m-1} f_{mn}\Bigg) + 
    \sum\limits_{m=1}^k  f_m\Bigg(\frac{r_m^{\mathcal{C}_b}}{r_m^{\mathcal{C}_t}}\Bigg) + 
    \sum\limits_{m=2}^k\sum\limits_{n=1}^{m-1} f_{mn}\Bigg(\frac{r_m^{\mathcal{C}_b}}{r_m^{\mathcal{C}_t}}\frac{r_n^{\mathcal{C}_b}}{r_n^{\mathcal{C}_t}}\Bigg)}
\end{multline}
In equation \ref{eqn:speedup_interactions}, the denominator of the right hand side represents the serial and parallel fractions contributing to the overall speedup. Specifically, $f_m$ represents the program fraction enhanced exclusively by the resource enhancement $\sfrac{r_m^{\mathcal{C}_b}}{r_m^{\mathcal{C}_t}}$ and $f_{mn}$ represents the program fraction enhanced by combined resource enhancement $\sfrac{r_m^{\mathcal{C}_b}r_n^{\mathcal{C}_b}}{r_m^{\mathcal{C}_t}r_n^{\mathcal{C}_t}}$. In this manner, up to two-factor interactions are captured in the speedup formulation.
\subsubsection{Enhancements due to higher order effects}
\label{ssec:enhancements_higherorder}
In some cases, it is desirable to include higher order interactions into the speedup equation. Consider the three term interaction $\mathcal{I}_{mnp} = (\sfrac{r_m^{\mathcal{C}_b}r_n^{\mathcal{C}_b}r_p^{\mathcal{C}_b}}{r_m^{\mathcal{C}_t}r_n^{\mathcal{C}_t}r_p^{\mathcal{C}_t}})$ along with the corresponding parallelizable program fraction $f_{mnp}$. This effect can be incorporated by adding $f_{mnp}(\mathcal{I}_{mnp}-1)$ to the denominator. Similarly, one may also incorporate other nonlinearities into the speedup equation. For example, consider the resource term representing the maximum memory bandwidth available per core. This term represents a nonlinear higher order interaction effect consisting of several factors such as channel width, channel count, memory frequency, number of cores, and core frequency. In this manner, the overall speedup equation can accommodate scaling across multiple resources and relevant nonlinear interactions.
\subsection{Amdahl's Law for data-driven modeling}
\label{sec:exp_pred_modeling}
The postulated speedup equations in Section \ref{ssec:amdahl_interactions} result in analytical models, which, while theoretically reasonable, require data for validation. The data offers insights into the contributions of the various resources and interactions for purposes of prediction. Therefore, Amdahl's law helps in hypothesizing a predictive model which is validated based on experimental data. Depending the predictive power of a given model, its complexity may be increased or reduced via the resource terms to avoid underfitting or overfitting \cite{james2013introduction}. This methodology results in a hybrid approach where both architectural knowledge and statistical techniques codetermine a multivariable analytical model. Further discussion on analytical modeling is provided in sections \ref{ssec:expt_design_modeling} and \ref{sec:modeling}.
\section{Experiment Setup}
\label{sec:experiment}
The efficacy of the analytical models in Section \ref{sec:methodology} is determined by using data from designed experiments. For this work, we conducted four experiments involving the \emph{SPECCPU 2017} and \emph{PCMark 10} benchmarks executed on Intel and AMD platforms. Let these experiments be represented by $\mathbf{E}_q \forall\ q \in \{1,2,3,4\}$. For each experiment, the benchmark and hardware platform combinations are shown in Table \ref{table:experiment}.
\begin{table}[H]
\centering
\renewcommand{\arraystretch}{1.2}
\resizebox{\textwidth}{!}{%
\begin{tabular}{ |c|c|c|c|c| }
 \hline
 \textbf{Experiment ID} & \textbf{Benchmark} & \textbf{Hardware} & \textbf{Operating System} & \textbf{Compiler} \\
 \hline
 $\mathbf{E}_1$ & SPECCPU 2017 & Intel Xeon 8180M & RHEL 7.3 & ICC 19u4 \\
 \hline
 $\mathbf{E}_2$ & SPECCPU 2017 & AMD EPYC 7702P & Ubuntu 19.04 & GCC 8.2 \\
 \hline
 $\mathbf{E}_3$ & PCMark 10 & Intel CoffeeLake 8700K & Win 10 Enterprise 1252 & N/A (Pre-compiled) \\
 \hline
 $\mathbf{E}_4$ & PCMark 10 & AMD Ryzen 3900X & Win 10 Pro 1903 & N/A (Pre-compiled) \\
 \hline
\end{tabular}}
\caption{Overview of Experiments}
\label{table:experiment}
\end{table}
\subsection{Workload settings}
\label{ssec:workload_settings}
For each benchmark, the settings employed during experimentation are described below:
\subsubsection{SPECCPU 2017}
\label{ssec:speccpu2017}
The SPECCPU 2017 benchmark consists of industry-standardized, CPU intensive suites for determining compute performance by stressing a system's processor, memory subsystem, and compiler \cite{bucek2018spec}. Among these suites, the \emph{SPECrate 2017 Integer} and \emph{SPECrate 2017 Floating Point} suites were both used to evaluate the performance of the \emph{Intel Xeon 8180M} and \emph{AMD EPYC 7702P} systems. The operational settings for both systems are depicted in the \textbf{Operating System} and \textbf{Compiler} columns of Table \ref{table:experiment}.
\subsubsection{PCMark 10}
\label{ssec:pcmark10}
The PCMark 10 benchmark represents a wide range of office activities from everyday productivity tasks to taxing work with digital media content \cite{pcmark10}. This benchmark is divided into groups each of which consists of several tests. In this work, the \emph{Essentials}, \emph{Productivity}, and \emph{Digital Content Creation} groups were all used to evaluate the perform of \emph{Intel CoffeeLake 8700K} and \emph{AMD Ryzen 3900X} systems. The operational settings for both systems are depicted in the \textbf{Operating System} and \textbf{Compiler} columns of Table \ref{table:experiment}.

\vspace{0.1in}
For each experiment $\mathbf{E}_q$, let the number of runs of the benchmark be denoted by $m_q$. Each run results in a score corresponding to a system with a specific configuration. Throughout each experiment, the benchmark and operational settings were fixed while the system configurations were varied.
\subsection{System configuration}
\label{ssec:system_configuration}
The resources provided by the system enable the execution of the benchmark. However, not all resources are varied in the course of the experiment. In this work, we employ the term \emph{resource variables} to refer only to the resources varied during the experiment. These variables are used to model the relationship between the system configuration and the benchmark performance. In general, any resource may be varied during the experiment to determine benchmark sensitivity. \added{In the conventional sense of the Amdahl's law, the benchmark sensitivity is quantified by the speedup due to the single resource enhancement. Based on sections \ref{ssec:need_to_extend_amdahls_law} and \ref{ssec:amdahl_multiple}, we attempt to quantify benchmark sensitivity as the speedup due to enhancement of mutually exclusive resources.} Here, we focus on exploring the speedup due to enhancing the CPU and memory subsystems and accordingly the resource variables include one or more aspects of the CPU such as the number of cores, the number of threads, core frequency, uncore frequency, and last level cache size; and one or more aspects of the memory subsystem such as the memory frequency and number of memory channels. Since the same resource variables support the execution of different workloads across various systems, the chosen variables generalize across systems and workloads. For each of the experiments undertaken, Table \ref{table:expt_config_ranges} depicts the range of configurations\footnote{In order to test the adaptability of the model across diverse design spaces, each experiment involved a subset of the possible variables.} utilized\deleted{:}\added{.}
\begin{table}[H]
\centering
\renewcommand{\arraystretch}{1.1}
\begin{tabular}{ |c|c|c|c|c| }
 \hline
 \textbf{\added{Resource} attribute \deleted{considered}} & $\mathbf{E}_1$ & $\mathbf{E}_2$ & $\mathbf{E}_3$ & $\mathbf{E}_4$ \\
 \hline
 \#DataPoints ($m_q$) & 58 & 20 & 177 & 38 \\ 
 \hline
 \#Cores (Min) & 1 & 64 & 1 & 1 \\ 
 \hline
 \#Cores (Max) & 28 & 64 & 6 & 12 \\
 \hline
 Core Freq. in MHz (Min) & 1800 & 1500 & 1200 & 2200 \\
 \hline
 Core Freq. in MHz (Max) & 2500 & 2000 & 3200 & 3800 \\
 \hline
 Uncore Freq. in MHz (Min) & 2200 & 1500 & 1200 & 2200 \\
 \hline
 Uncore Freq. in MHz (Max) & 2200 & 2000 & 32   00 & 3800 \\
 \hline
 LLC MB (Min) & 7 & 128 & 3 & 64 \\
 \hline
 LLC MB (Max) & 38 & 256 & 12 & 64 \\
 \hline
 Mem Freq. in MHz (Min) & 2133 & 2666 & 1300 & 1333 \\
 \hline
 Mem Freq. in MHz (Max) & 2667 & 3200 & 2667 & 1600 \\
 \hline
 \#MemCH (Min) & 6 & 8 & 1 & 4 \\
 \hline
 \#MemCH (Max) & 6 & 8 & 2 & 4 \\
 \hline
\end{tabular}
\caption{Range of the Experimental Design Spaces}
\label{table:expt_config_ranges}
\end{table}
In addition to the above resource variables, the simultaneous multithreading state (ON or OFF) was varied during the experiment.
\subsection{Relation between experimental design and modeling}
\label{ssec:expt_design_modeling}
The relation between experiment design and modeling is crucial in developing analytical models with appropriate modeling assumptions. The number or set of points required to be collected is driven by the underlying purpose of the experiment. In most cases, the purpose is to accurately model the predictive relationship between the system resources and the benchmark output (score).

\emph{Explanatory and predictive modeling:} It is important to note that predictive models do not necessarily reflect the true underlying relationship. For example, the relation between system resources and benchmark scores is dynamic in nature and hence simulated in time whereas most analytical models are static. Therefore, analytical modeling is not meant to provide explanatory or "true" models of the underlying relationship but only to generate predictions based on simplified models. However, this does not necessarily imply analytical or predictive models are less accurate as they have shown to even outperform explanatory models in some cases \cite{shmueli2010explain}. This counter-intuitive behavior is possible due to reasons such as small magnitude of model parameters, noisy data, regressor correlations, or small sample sizes \cite{shmueli2010explain}.

\emph{Experiment design for analytical modeling:} \added{The experiment design involves selecting the design space used for analytical modeling. The functional form of the analytical model is selected to represent the underlying data-generating process.} Upon specifying the analytical model, the design space spanned by the resource variables, along with the corresponding benchmark scores, are used for the estimation procedure. \added{The estimation procedure has a bearing on the set of points required \cite{kramer1980finite}. In this work, we employ least squares estimation.} \deleted{In general, different estimation procedures require different number of data points for estimating model parameters. The conventional Amdahl's law formulation (Equation \ref{eqn:speedup_i}) examines speedup due to a single resource or One Factor At a Time (OFAT) and requires a minimum of three data points to estimate the parameters. However, this number varies based on the predictability of the relationship benchmark score and resource variables in light of the model considered. Here, we focus on building general multivariable models based on the extended Amdahl's law (Equation \ref{eqn:speedup_interactions}) and the number of data points gathered for each experiment is provided in Table \ref{table:expt_config_ranges}.} \added{In general, the set of points required to obtain acceptable accuracy varies based on the efficiency of the estimator. While the investigation of the optimal design space \cite{goos2011optimal} is beyond the scope of the present work, we nevertheless examine diverse design spaces to assess the efficacy of the extended Amdahl's law (Equation \ref{eqn:speedup_interactions}) per the design scope shown in Table \ref{table:expt_config_ranges}. While any random sampling or systematic design of experiments-based approaches could be employed within the feasible set of configurations, we used random sampling based on a One-Factor-At-a-Time (OFAT) approach to generate diverse design spaces. The idea behind employing diverse design spaces is to demonstrate the generalizability of the proposed extension.} In what follows, we describe the procedure for model design, training, and validation.
\section{Model design, training, and validation}
\label{sec:modeling}
The data obtained from an experiment $\mathbf{E}_q$ consists of $m_q$ configurations and scores for fixed values of the benchmark and operational settings. From Section \ref{sec:methodology}, it may be noted that each configuration $\mathcal{C}$ is made up of one or more resource variables $r_i^\mathcal{C}$ which are varied across runs to obtain the corresponding scores. 
\subsection{Analytical modeling}
Analytical models represent a predictive relationship between the independent system resources and the dependent target variable\deleted{s, often trivially related to the score}. Accordingly, the target and resource variables need to be specified and an underlying predictive relationship needs to \added{be} determined and validated using experimental data.
\subsubsection{Target variable}
The target variable \added{is the response such as the benchmark score associated with}\deleted{refers to the variable that is intended to be predicted by} the model. If higher scores are considered better they \deleted{represent}\added{are equivalent to} the speedup\deleted{equivalent}. Similarly, if lower scores are considered better, they \deleted{represent}\added{are} equivalent to the inverse speedup. Here, the target variable is equivalent to the inverse speedup as detailed in Section \ref{sssec:modeling}. \added{This is because}\deleted{For example, in case of} \emph{SPECCPU 2017} SpecRate \added{and PCMark 10}\deleted{the} \added{output} scores \deleted{output by the benchmark are}proportional to the speedup \cite{bucek2018spec} and hence their inverse \added{score}\deleted{score is chosen to be}\added{is} the target variable. \deleted{In case of \emph{PCMark 10} the higher the output score the better the system under test. Thus, these scores maybe viewed as equivalents of speedups and hence the inverse scores are chosen to be the target variable.}
\subsubsection{Resource variables}
The value of the target variable is dependent on the resource variables. This dependency is specified by the model. The set of resource variables identified in Section \ref{ssec:system_configuration} are incorporated into the model as model features\footnote{$X_{k,i}$ represents the features of the model described in Equation \ref{eqn:amhdal_regression_model}.}. We note that each feature may either be generic as depicted in equations \ref{eqn:speedup_m}, \ref{eqn:speedup_interactions} or engineered using knowledge of the architecture and/or the benchmark execution. In this work, we use both types of features in the model. The generic features represent the independent variables such as frequencies and core counts. The engineered features used represent meaningful resources such as cache size available per core or the maximum bandwidth per core noted in Section \ref{ssec:enhancements_higherorder}.
\subsubsection{Designing Amdahl's law-based regression models}
\label{sssec:modeling}
Given the target and resource variable specifications, the explanatory relationship between these variables is provided for by the extended Amdahl's law-based speedup equations similar to the forms shown in equations \ref{eqn:speedup_m} or \ref{eqn:speedup_interactions}. Each of these equations offer varying degrees of predictive ability quantifiable by experimental data. As noted in Section \ref{sec:exp_pred_modeling}, the explanatory relationship offered by equations such as \ref{eqn:speedup_m}, \ref{eqn:speedup_interactions} is leveraged to build analytical models. Consider equation \ref{eqn:speedup_interactions}, wherein we rewrite the denominator after making the following substitutions. Let $\scriptstyle\alpha_0\ \coloneqq\ (1-\sum\limits_{m=1}^k f_m-\sum\limits_{m=2}^k\sum\limits_{n=1}^{m-1} f_{mn})$, $\scriptstyle\alpha_p\ \coloneqq\ f_p\ \forall\ p\ \in \{1,\cdots,k\}$, and $\scriptstyle\alpha_q\ \coloneqq\ f_{mn}\ \forall\ m\ \in\ \{1,\cdots,k\},\ n\ \in\ \{1,\cdots,m-1\},\ q\ \in\ \{k+1,\cdots,\mathcal{K}\}$, where $\mathcal{K}=\Mycomb[k]{1}+\Mycomb[k]{2}$ denotes the total number of single factors and two-factor interactions in the model. Further, let $\scriptstyle\tilde{X}_0\ \coloneqq\ 1$, $\scriptstyle\tilde{X}_p\ \coloneqq\ (\sfrac{r_p^{\mathcal{C}_t}}{r_p^{\mathcal{C}_b}})\ \forall\ p\ \in \{1,\cdots,k\}$ and $\scriptstyle\tilde{X}_q\ \coloneqq\ (\sfrac{r_m^{\mathcal{C}_t}r_n^{\mathcal{C}_t}}{r_m^{\mathcal{C}_b}r_n^{\mathcal{C}_b}})\ \forall\ m\ \in\ \{1,\cdots,k\},\ n\ \in\ \{1,\cdots,m-1\},\ q\ \in\ \{k+1,\cdots,\mathcal{K}\}$. Accordingly, the denominator can be rewritten as $\scriptstyle\sum\limits_{k=0}^\mathcal{K} \sfrac{\alpha_k}{\tilde{X}_k}$, encapsulating the resource variables, interactions, and parallel fraction-based coefficients. 
The left side represents the performance speedup $\sfrac{P^{\mathcal{C}_t}}{P^{\mathcal{C}_b}}$. We note that the speedup equation applies to all runs of an experiment. Extending the above convention for each run $i$, let the speedup equivalent be denoted by $\tilde{Y}_i$, the $k^{th}$ resource variable/interaction term by $\tilde{X}_{k,i}$, and $\tilde{X}_0$ by $\tilde{X}_{0,i}$. For convenience, we will refer to all the terms $\tilde{X}_{k,i}$ including $\tilde{X}_{0,i}$ as resource terms. Thus, the speedup equations in Section \ref{ssec:amdahl_interactions} corresponding to the $i^{th}$ run can be represented as:
\begin{equation}
\label{eqn:amdahl_to_regression}
    \tilde{Y}_i = \frac{1}{\sum\limits_{k=0}^\mathcal{K}\alpha_k \frac{1}{\tilde{X}_{k,i}}}
\end{equation}
\\
where, $\alpha_k$ is the refactored coefficient of the $k^{th}$ resource term and $\mathcal{K}$ denotes the number of resource terms considered for modeling. It is easy to see that the above equation may be readily recast into a linear regression model suitable for learning:
\begin{equation}
\label{eqn:amhdal_regression_model}
    Y_i = \sum\limits_{k=0}^\mathcal{K}\alpha_k X_{k,i}
\end{equation}
where, $Y_i = \frac{1}{\tilde{Y}_i}$ and $X_i = \frac{1}{\tilde{X}_i}$ are the reciprocal transformations used to recast equation \ref{eqn:amdahl_to_regression}. In this manner, analytical models may be designed based on Amdahl's law and regression techniques. 
\subsubsection{Model training and cross-validation}
\label{sssec:cross_validation}
Given a model design, its feature weights (coefficients) are determined by training the model. These coefficients specify a trained model, also known as a model instance. In this work, the models are trained using ordinary least squares implemented by the python library \emph{scikit-learn}. The features of the model were scaled via normalization during preprocessing to improve model training.

\begin{figure}[ht]
    \centering
    \includegraphics[width=0.75\textwidth]{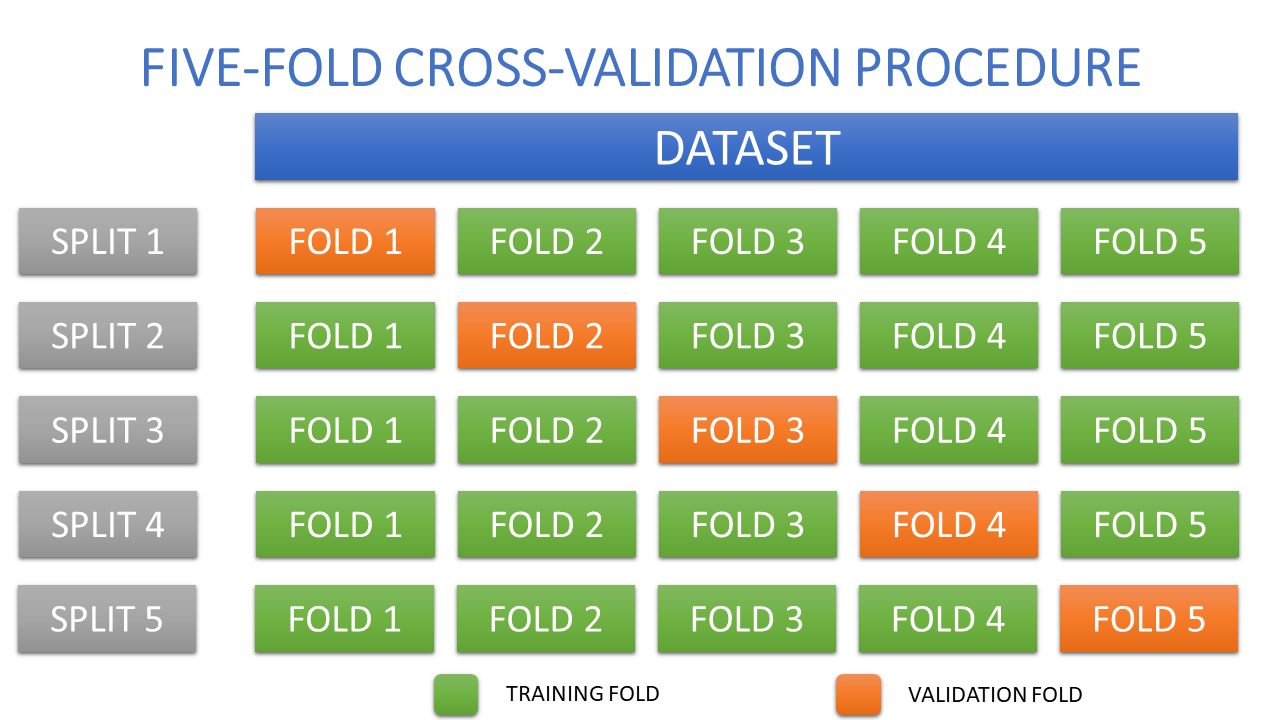}
    \caption{Five-fold cross-validation depicting training and validation folds}
    \label{FIG:cross_validation}
\end{figure}

The performance of a model needs to be evaluated to determine its validity for prediction purposes. We use \emph{Mean Absolute Percentage Error (MAPE)} as the model evaluation metric. The validation procedure consists of evaluating the model on data not used for training. In this work, we perform model validation by \emph{five-fold cross-validation} as it provides an accurate estimate of the expected generalization error and uses the data efficiently for validation \cite{james2013introduction}. The measurements dataset consisting of $m_q$ observations is split into five parts called folds. Each split uses four parts for training and one part for validation as shown in Figure \ref{FIG:cross_validation}. Thus, each split reserves 80\% of the data for training and 20\% for validation testing. The validation errors across all folds are averaged to determine the expected generalization error of the model.

\begin{figure}[ht]
    \centering
    \includegraphics[width=0.75\textwidth]{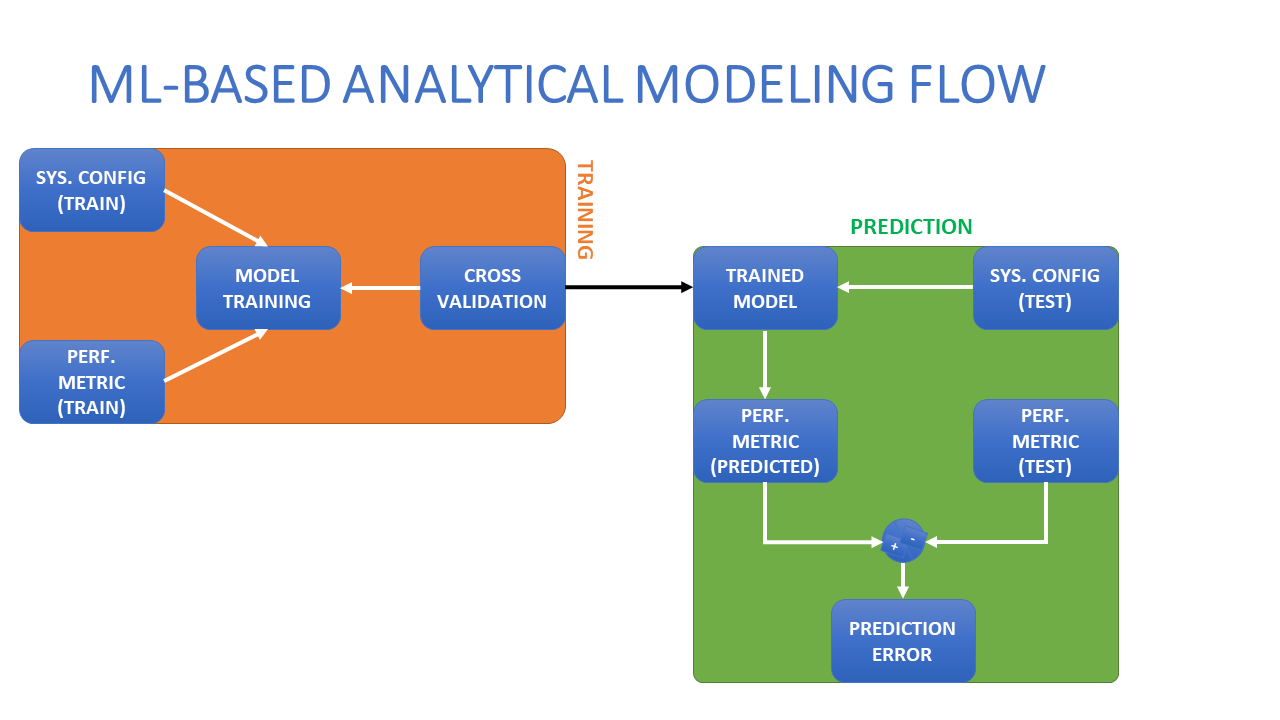}
    \caption{Analytical modeling within a Machine Learning (ML) framework}
    \label{FIG:analytical_modeling}
\end{figure}

Once the model is cross-validated and the expected generalization errors are deemed satisfactory, it is used for prediction. The end-to-end analytical modeling flow consisting of the training, validation, and prediction phases are depicted in Figure \ref{FIG:analytical_modeling}.
\section{Results and discussion}
\label{sec:results}
For each of the four experiments $\mathbf{E}_1, \mathbf{E}_2, \mathbf{E}_3,$ and $\mathbf{E}_4$, multi-variable regression models based on extended Amdahl's law were developed and cross-validated as described in Section \ref{sec:modeling}. The cross-validation results evaluate the effectiveness of the model for purposes of prediction. As stated in Section \ref{sssec:cross_validation} we use the \emph{Mean Absolute Percentage Error} (MAPE) as the evaluation metric. For depicting the results below, we convert MAPE into the the equivalent accuracy by calculating \emph{100-MAPE}. The average of the \emph{Mean Absolute Percentage Accuracy} for all validation folds is shown in the accuracy results below for each benchmark and hardware platform.
\subsection{Experiments involving SPECCPU 2017}
These experiments involved executing the SpecRate Integer and Floating point suites within the SPECCPU2017 benchmark. The Integer suite consists of 11 tests and the Floating Point suite consists of 14 tests. Twenty five models were constructed and cross-validated corresponding to each of the 25 tests spanning both suites.
\subsubsection{$\mathbf{E}_1$: SPECCPU 2017 on Intel Xeon 8180M}
\label{ssec:spec_intel_results}
\begin{figure}
\centering
\begin{subfigure}{0.6\linewidth}
    \centering
    \includegraphics[width=\linewidth]{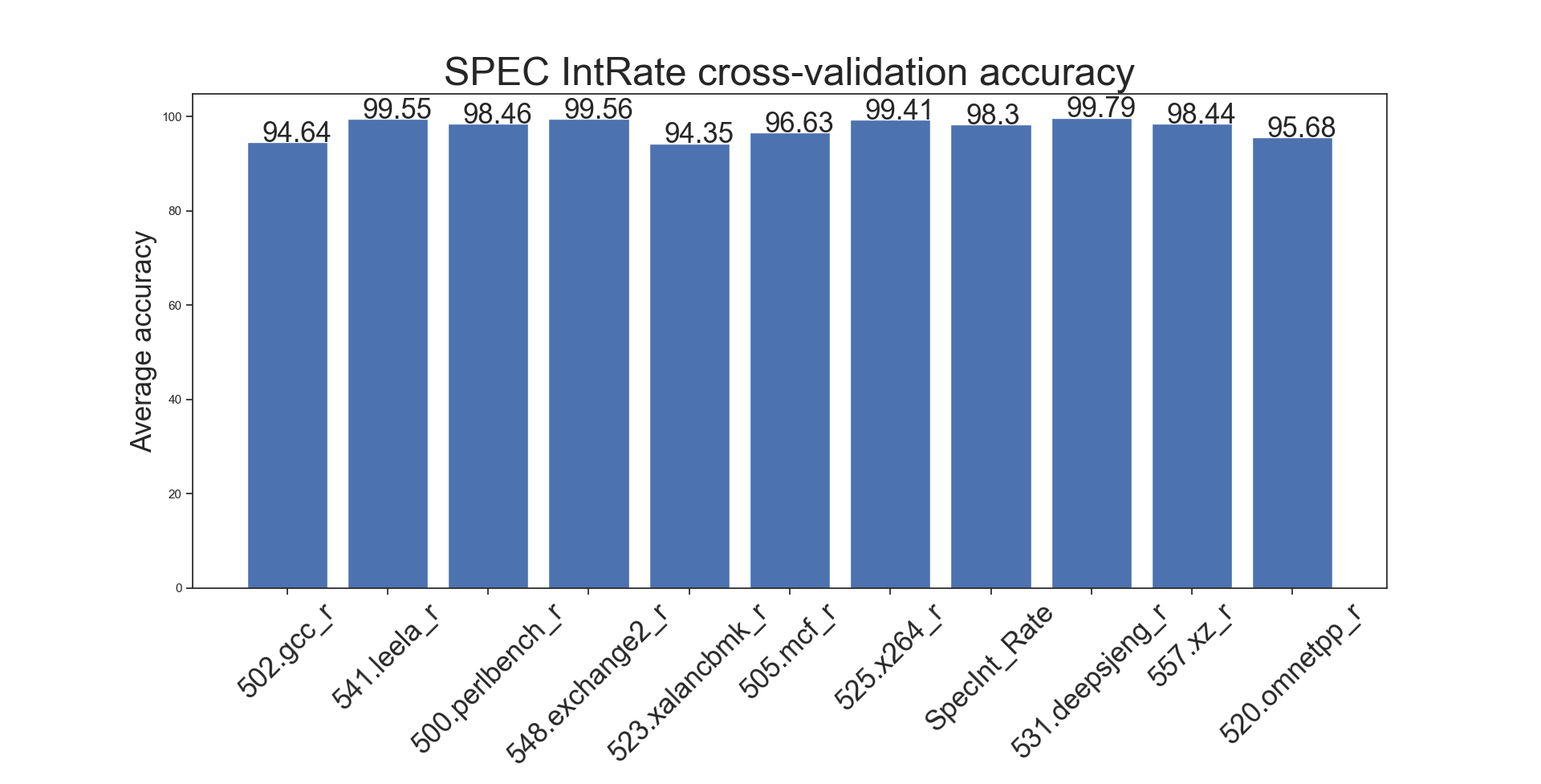}
    \caption{SpecRate Integer}
    \label{FIG:sir_intel_results}
\end{subfigure}%
\begin{subfigure}{0.6\linewidth}
    \centering
    \includegraphics[width=\linewidth]{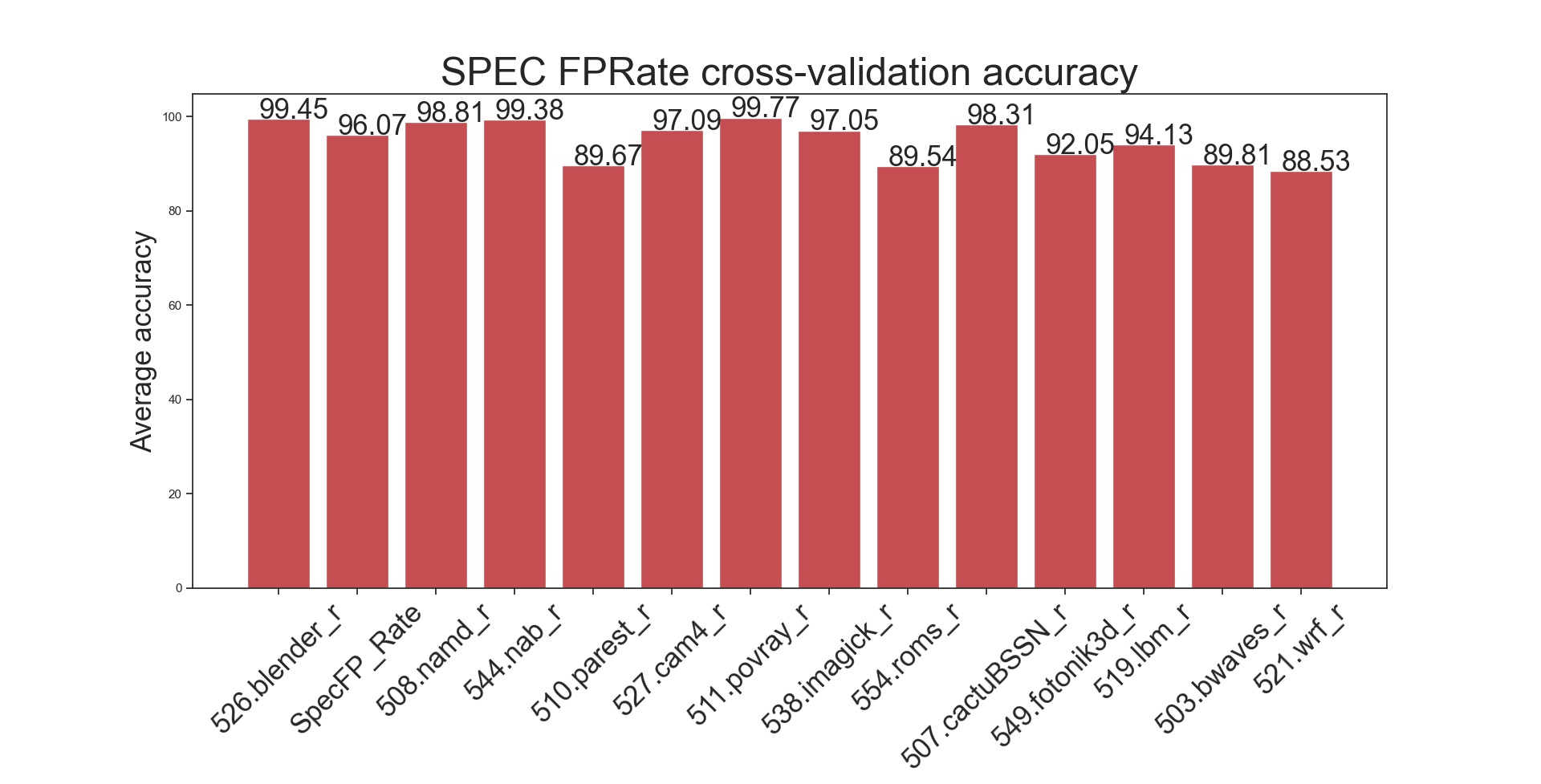}
    \caption{SpecRate Floating Point}
    \label{FIG:sfr_intel_results}
\end{subfigure}
\caption{SPEC CPU 2017 cross-validation results on Intel Xeon 8180M}
\label{FIG:spec_intel_results}
\end{figure}

Figures \ref{FIG:sir_intel_results} and \ref{FIG:sfr_intel_results} depict the cross-validation accuracy of the tests in the Integer and Floating Point suites benchmarked on Intel Xeon 8180M. In case of the Integer \added{suite}\deleted{tests}, the \added{underlying} models result in an accuracy of $\approx 95\%$ or higher. However, in case of the Floating point tests, the \emph{wrf\_r} model resulted in the minimum accuracy of $\approx 88\%$ and some models for other tests resulted in the maximum accuracy of $> 99\%$. The overall models for the Integer and Floating point suites resulted in average accuracies of $98\%$ and $96\%$, respectively.
\subsubsection{$\mathbf{E}_2$: SPECCPU 2017 on AMD EYPC 7702P}
\label{ssec:spec_amd_results}
\begin{figure}
\centering
\begin{subfigure}{0.6\linewidth}
    \centering
    \includegraphics[width=\linewidth]{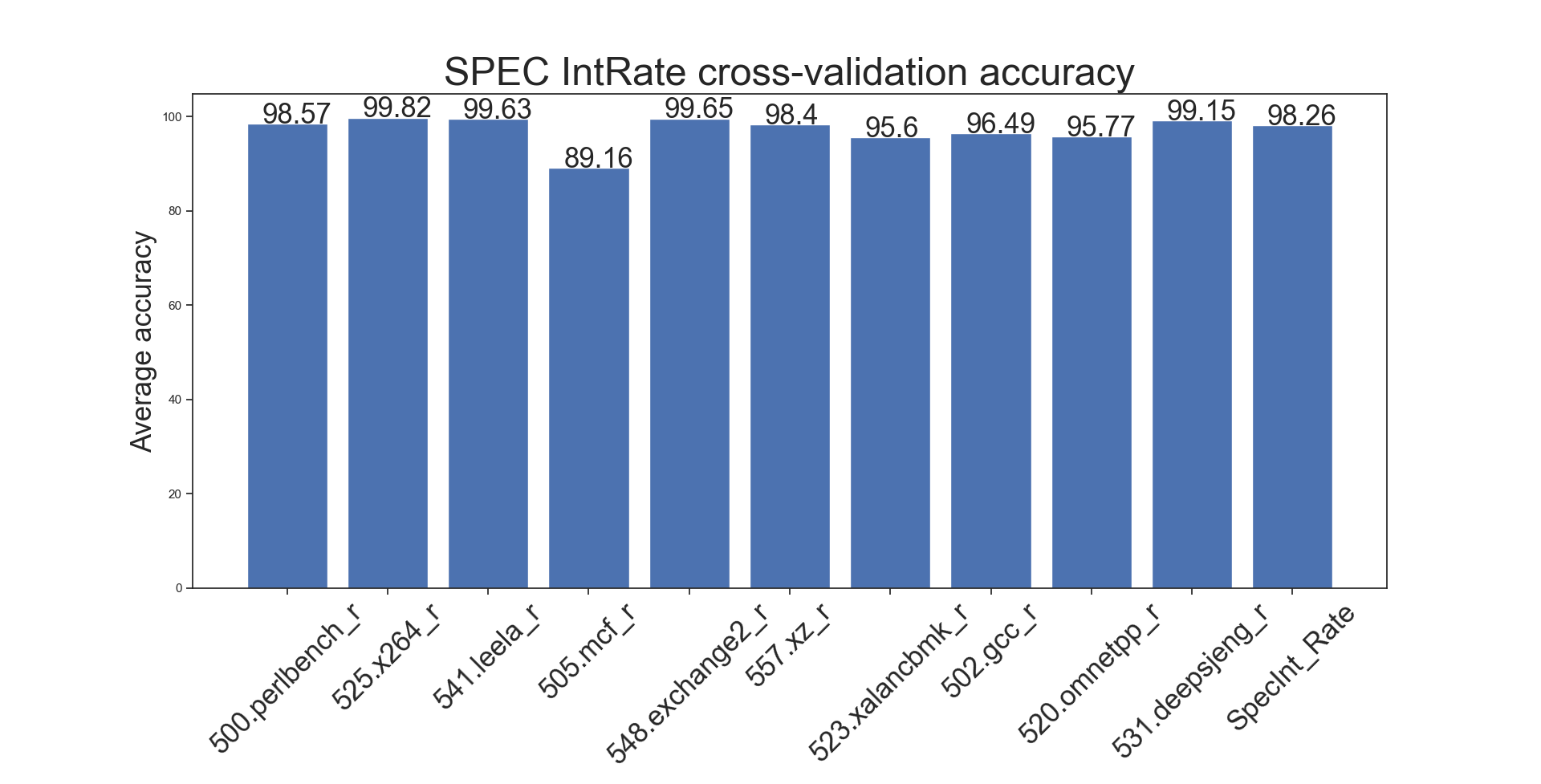}
    \caption{SpecRate Integer}
    \label{FIG:sir_amd_results}
\end{subfigure}%
\begin{subfigure}{0.6\linewidth}
    \centering
    \includegraphics[width=\linewidth]{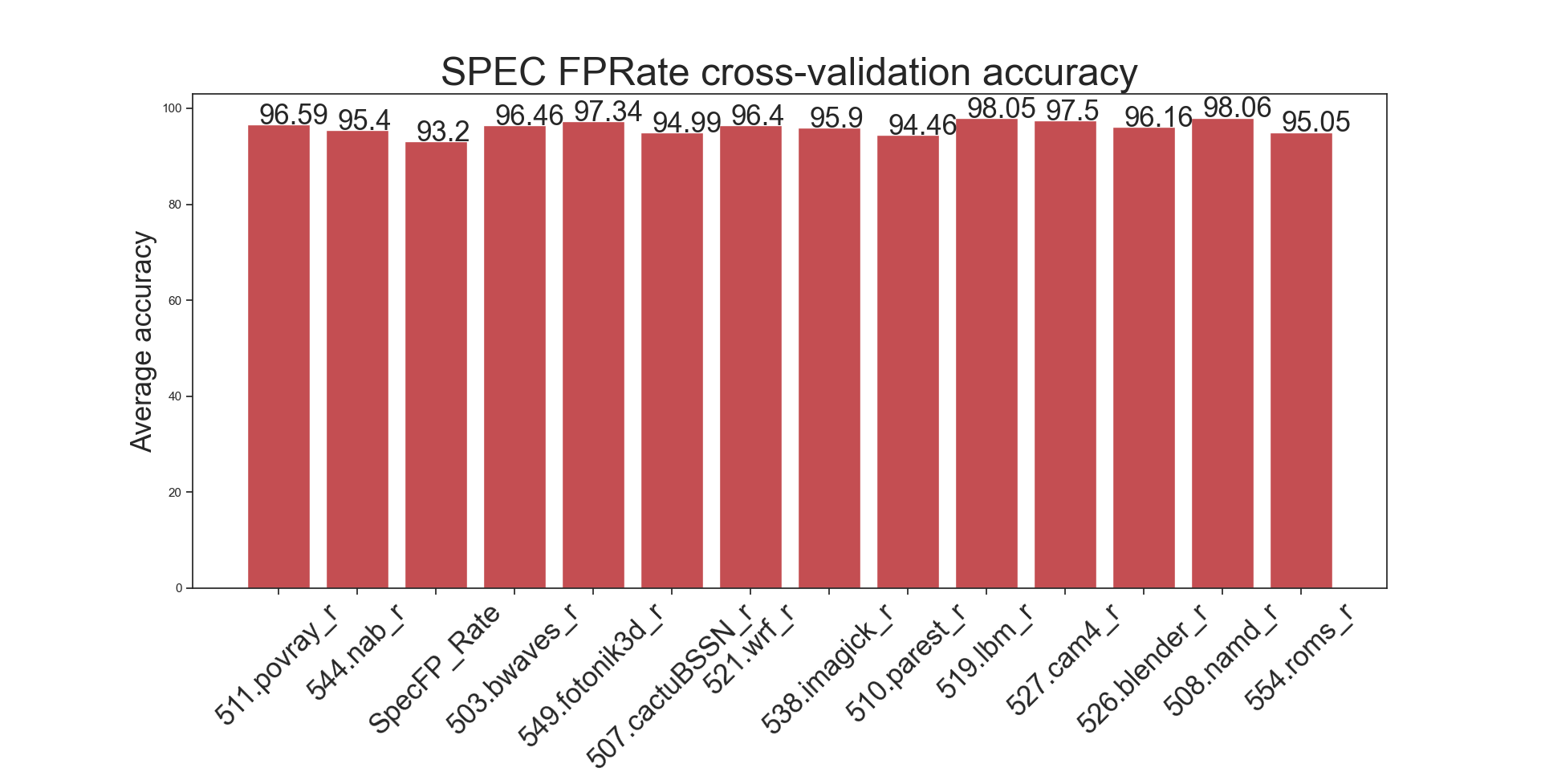}
    \caption{SpecRate Floating Point}
    \label{FIG:sfr_amd_results}
\end{subfigure}
\caption{SPEC CPU 2017 cross-validation results on AMD EPYC 7702P}
\label{FIG:spec_amd_results}
\end{figure}

Figures \ref{FIG:sir_amd_results} and \ref{FIG:sfr_amd_results} depict the cross-validation accuracy of the tests in the Integer and Floating Point suites benchmarked on AMD EYPC 7702P. In case of the Integer tests, model accuracies are greater than $95\%$ with the exception of \emph{mcf\_r} whose model accuracy was $\approx 89\%$. In the Floating point tests, most tests resulted in $\approx 95\%$ accuracy or greater. The overall models for Integer and Floating point suites resulted in $98\%$ and $93\%$, respectively.
\subsection{Experiments involving PCMark 10}
The experiments involved executing the PCMark 10 benchmark on each of the platforms below. The benchmark consists of three test groups namely, Essentials, Productivity, and Digital Content Creation. Each of test groups have several tests. For demonstrating model efficacy, we examine the cross-validated accuracies of four models each corresponding to \added{either} one of the three test groups or \added{to the aggregate output of} the \deleted{overall} PCMark 10 benchmark.
\subsection{$\mathbf{E}_3$: PCMark 10 on Intel CoffeeLake 8700K}
\label{ssec:pcmark_intel_results}
\begin{figure}
\centering
\begin{subfigure}{0.6\linewidth}
    \centering
    \includegraphics[width=\linewidth]{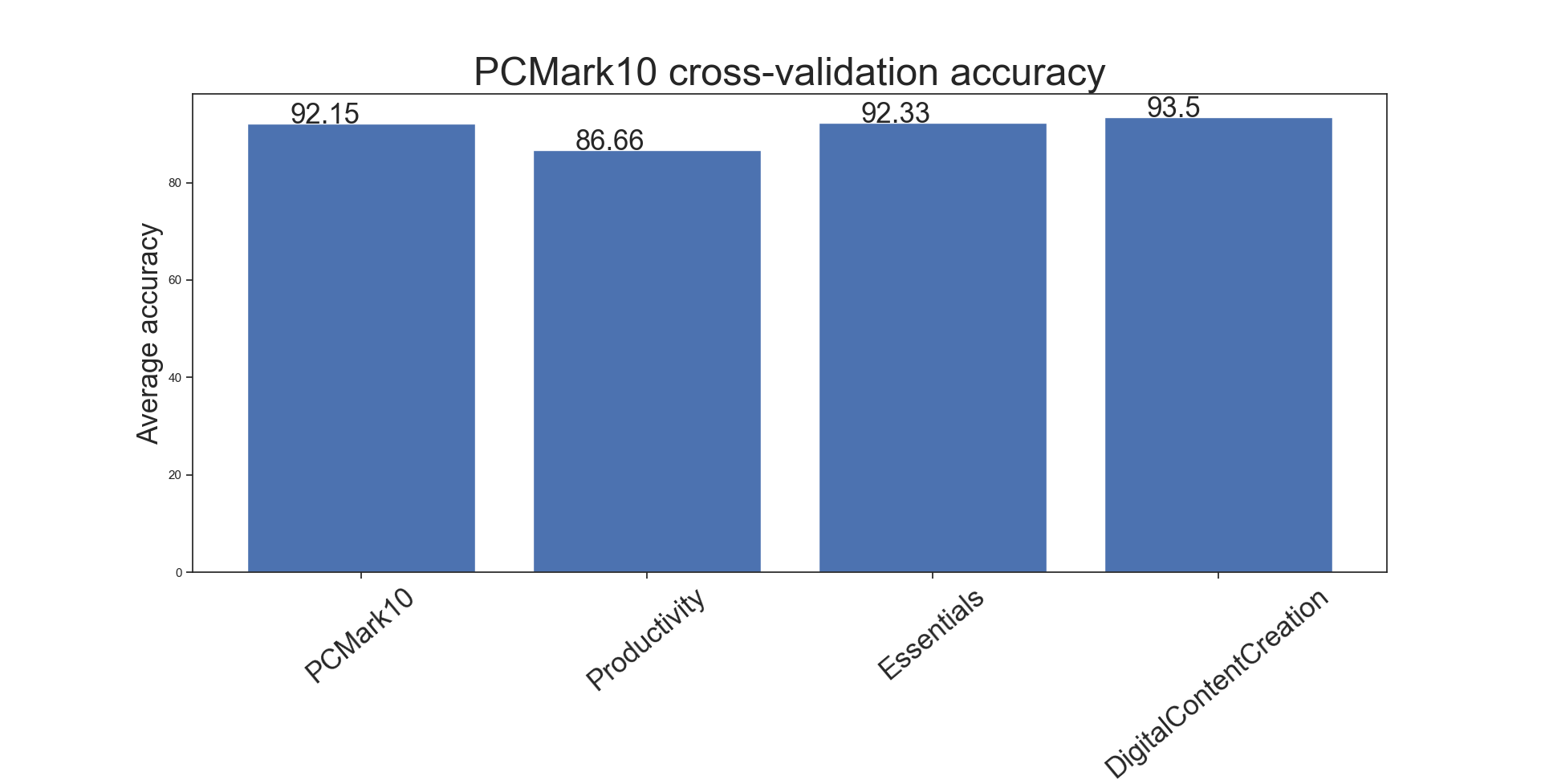}
    \caption{PCMark10 on Intel CoffeeLake 8700K}
    \label{FIG:pcmark_intel_results}
\end{subfigure}%
\begin{subfigure}{0.6\linewidth}
    \centering
    \includegraphics[width=\linewidth]{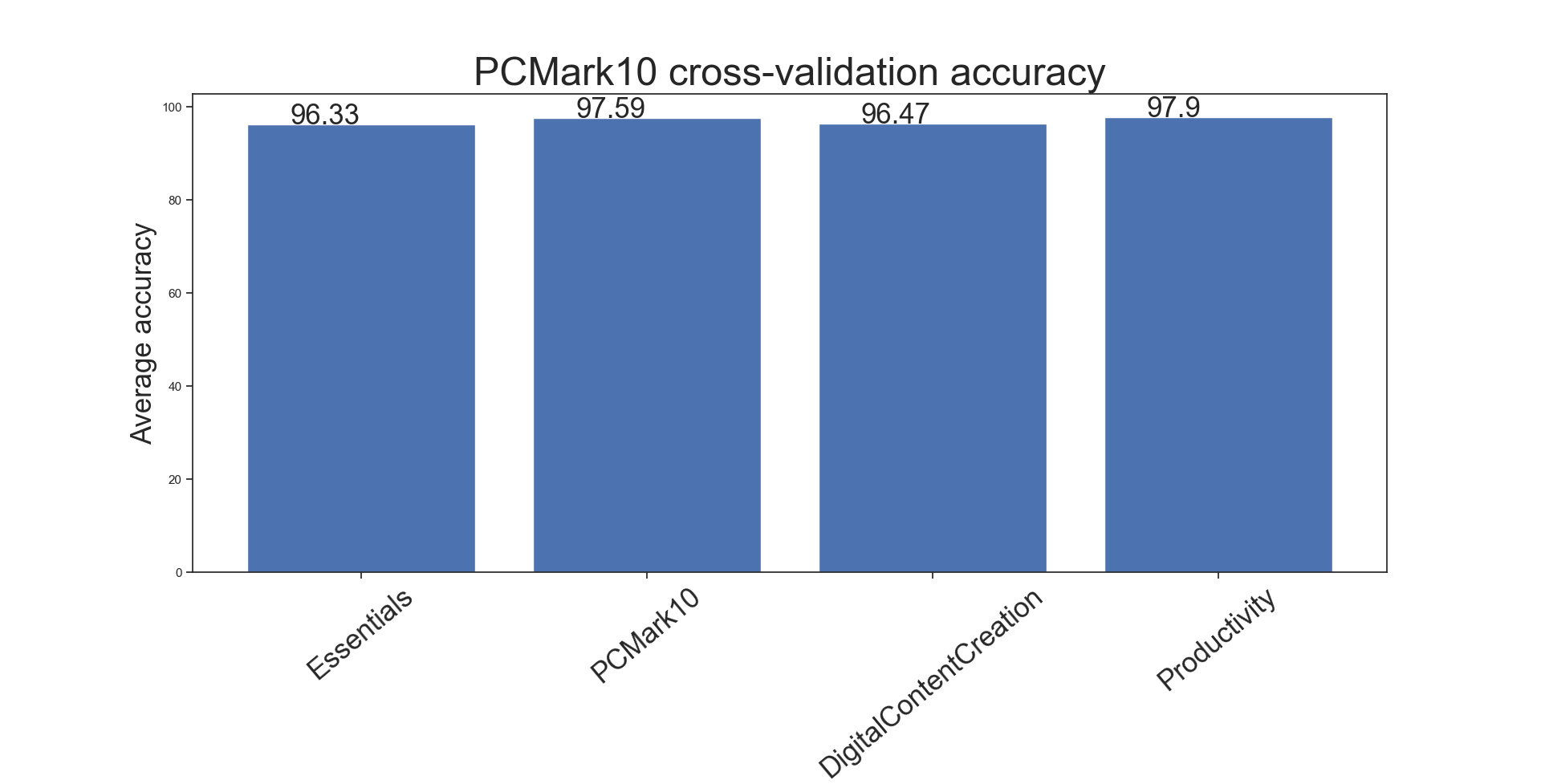}
    \caption{PCMark10 on AMD Ryzen 3900X}
    \label{FIG:pcmark_amd_results}
\end{subfigure}
\caption{PCMark10 on Intel CFL 8700K \& AMD Ryzen 3900X: Cross-validation results}
\label{example}
\end{figure}

Figure \ref{FIG:pcmark_intel_results} depicts the cross-validation accuracy of the models corresponding to each of the three test groups in PCMark 10. The accuracies of the Essentials, Productivity, and Digital Content Creation models are found to be $92\%$, $88\%$, and $92\%$, respectively. The overall PCMark 10 model results in an average accuracy of $92\%$. 
\subsection{$\mathbf{E}_4$: PCMark 10 on AMD Ryzen 3900X}
\label{ssec:pcmark_amd_results}
Figure \ref{FIG:pcmark_amd_results} depicts the cross-validation accuracy of the models corresponding to each of the three test groups in PCMark 10. The accuracies of the Essentials, Productivity, and Digital Content Creation models are found to be $96\%$, $98\%$, and $96\%$, respectively. The overall PCMark 10 model results in an accuracy of $98\%$. We note that the model accuracies on the Ryzen 3900X are found to be higher than the respective model accuracies on the CoffeeLake 8700K. This indicates model learning varies according to the platform architecture for a given benchmark.
\subsection{\added{Results of cross-validation}}
\label{ssec:results_cross_validation}
\added{The models corresponding to the fifty-eight (58) tests were cross-validated per the five-fold procedure outlined in Section \ref{sssec:cross_validation}. The cross-validation accuracies for a few tests are provided in Table \ref{table:cross_validation_accuracy} below.}
\begin{table}[H]
\centering
\renewcommand{\arraystretch}{1.2}
\resizebox{\textwidth}{!}{%
\begin{tabular}{ |c|c|c|c|c|c|c| }
 \hline
 \textbf{Experiment} & \textbf{Test} & Fold 1 & Fold 2 & Fold 3 & Fold 4 & Fold 5 \\
 \hline
 $\mathbf{E}_1$ & SpecInt\_Rate & 96.68\% & 98.85\% & 99.01\% & 98.40\% & 98.61\% \\
 \hline
 $\mathbf{E}_2$ & SpecFP\_Rate & 79.74\% & 94.63\% & 97.60\% & 98.40\% & 95.61\% \\
 \hline
 $\mathbf{E}_3$ & Productivity & 92.78\% & 83.94\% & 82.36\% & 91.09\% & 87.83\% \\
 \hline
 $\mathbf{E}_4$ & PCMark10 (Overall) & 97.83\% & 99.22\% & 99.59\% & 91.44\% & 99.33\% \\
 \hline
\end{tabular}}
\caption{Mean Absolute Percentage Accuracy across the cross-validation folds}
\label{table:cross_validation_accuracy}
\end{table}
The configurations used in the cross-validation procedure (Figure \ref{FIG:cross_validation}) corresponding to experiment $\mathbf{E}_1$ are listed in Table \ref{table:cross_validation_configs}.
\begin{table}[!ht]
    \centering
    \renewcommand{\arraystretch}{1.6}
    \resizebox{\textwidth}{!}{%
    \begin{tabular}{|l|l|l|l|l|l|l|l|l|l|}
    \hline
        Fold sequence & LLC (MB) & NumCores & Mem. Freq. (MHz) & NumMemCH & Core Freq. (MHz) & Uncore Freq. (MHz) & NumThreadsPerCore & LLC\_per\_core (derived) & B/W available per Core-Clk (derived)  \\ \hline
        VTTTT & 38 & 8 & 2667 & 6 & 2500 & 2200 & 2 & 4.75 & 0.00640080000000000  \\ \hline
        VTTTT & 38 & 12 & 2667 & 6 & 1800 & 2200 & 1 & 3.17 & 0.00592666666666667  \\ \hline
        VTTTT & 38 & 12 & 2667 & 6 & 2200 & 2200 & 1 & 3.17 & 0.00484909090909091  \\ \hline
        VTTTT & 38 & 12 & 2667 & 6 & 2500 & 2200 & 1 & 3.17 & 0.00426720000000000  \\ \hline
        VTTTT & 38 & 12 & 2667 & 6 & 1800 & 2200 & 2 & 3.17 & 0.00592666666666667  \\ \hline
        VTTTT & 38 & 12 & 2667 & 6 & 2200 & 2200 & 2 & 3.17 & 0.00484909090909091  \\ \hline
        VTTTT & 38 & 12 & 2667 & 6 & 2500 & 2200 & 2 & 3.17 & 0.00426720000000000  \\ \hline
        VTTTT & 38 & 18 & 2667 & 6 & 2500 & 2200 & 1 & 2.11 & 0.00284480000000000  \\ \hline
        VTTTT & 38 & 18 & 2667 & 6 & 2500 & 2200 & 2 & 2.11 & 0.00284480000000000  \\ \hline
        VTTTT & 38 & 1 & 2667 & 6 & 2500 & 2200 & 1 & 38.00 & 0.05120640000000000  \\ \hline
        VTTTT & 38 & 1 & 2667 & 6 & 2500 & 2200 & 2 & 38.00 & 0.05120640000000000  \\ \hline
        VTTTT & 38 & 20 & 2667 & 6 & 1800 & 2200 & 1 & 1.90 & 0.00355600000000000  \\ \hline
        TVTTT & 38 & 20 & 2667 & 6 & 2200 & 2200 & 1 & 1.90 & 0.00290945454545454  \\ \hline
        TVTTT & 38 & 20 & 2667 & 6 & 2500 & 2200 & 1 & 1.90 & 0.00256032000000000  \\ \hline
        TVTTT & 38 & 20 & 2667 & 6 & 1800 & 2200 & 2 & 1.90 & 0.00355600000000000  \\ \hline
        TVTTT & 38 & 20 & 2667 & 6 & 2200 & 2200 & 2 & 1.90 & 0.00290945454545454  \\ \hline
        TVTTT & 38 & 20 & 2667 & 6 & 2500 & 2200 & 2 & 1.90 & 0.00256032000000000  \\ \hline
        TVTTT & 38 & 22 & 2667 & 6 & 2500 & 2200 & 1 & 1.73 & 0.00232756363636364  \\ \hline
        TVTTT & 38 & 22 & 2667 & 6 & 2500 & 2200 & 2 & 1.73 & 0.00232756363636364  \\ \hline
        TVTTT & 38 & 24 & 2667 & 6 & 2500 & 2200 & 1 & 1.58 & 0.00213360000000000  \\ \hline
        TVTTT & 38 & 24 & 2667 & 6 & 2500 & 2200 & 2 & 1.58 & 0.00213360000000000  \\ \hline
        TVTTT & 38 & 28 & 2667 & 6 & 1800 & 2200 & 1 & 1.36 & 0.00254000000000000  \\ \hline
        TVTTT & 38 & 28 & 2667 & 6 & 2000 & 2200 & 1 & 1.36 & 0.00228600000000000  \\ \hline
        TVTTT & 38 & 28 & 2667 & 6 & 2200 & 2200 & 1 & 1.36 & 0.00207818181818182  \\ \hline
        TTVTT & 38 & 28 & 2667 & 6 & 2500 & 2200 & 1 & 1.36 & 0.00182880000000000  \\ \hline
        TTVTT & 38 & 28 & 2667 & 6 & 1800 & 2200 & 2 & 1.36 & 0.00254000000000000  \\ \hline
        TTVTT & 38 & 28 & 2667 & 6 & 2000 & 2200 & 2 & 1.36 & 0.00228600000000000  \\ \hline
        TTVTT & 38 & 28 & 2667 & 6 & 2200 & 2200 & 2 & 1.36 & 0.00207818181818182  \\ \hline
        TTVTT & 38 & 28 & 2667 & 6 & 2500 & 2200 & 2 & 1.36 & 0.00182880000000000  \\ \hline
        TTVTT & 38 & 4 & 2667 & 6 & 2500 & 2200 & 1 & 9.50 & 0.01280160000000000  \\ \hline
        TTVTT & 38 & 4 & 2667 & 6 & 2500 & 2200 & 2 & 9.50 & 0.01280160000000000  \\ \hline
        TTVTT & 38 & 8 & 2667 & 6 & 2500 & 2200 & 1 & 4.75 & 0.00640080000000000  \\ \hline
        TTVTT & 14 & 28 & 2667 & 6 & 2500 & 2200 & 1 & 0.50 & 0.00182880000000000  \\ \hline
        TTVTT & 21 & 28 & 2667 & 6 & 2500 & 2200 & 1 & 0.75 & 0.00182880000000000  \\ \hline
        TTVTT & 28 & 28 & 2667 & 6 & 2500 & 2200 & 1 & 1.00 & 0.00182880000000000  \\ \hline
        TTVTT & 35 & 28 & 2667 & 6 & 2500 & 2200 & 1 & 1.25 & 0.00182880000000000  \\ \hline
        TTTVT & 7 & 28 & 2667 & 6 & 2500 & 2200 & 1 & 0.25 & 0.00182880000000000  \\ \hline
        TTTVT & 14 & 28 & 2667 & 6 & 2500 & 2200 & 2 & 0.50 & 0.00182880000000000  \\ \hline
        TTTVT & 21 & 28 & 2667 & 6 & 2500 & 2200 & 2 & 0.75 & 0.00182880000000000  \\ \hline
        TTTVT & 28 & 28 & 2667 & 6 & 2500 & 2200 & 2 & 1.00 & 0.00182880000000000  \\ \hline
        TTTVT & 35 & 28 & 2667 & 6 & 2500 & 2200 & 2 & 1.25 & 0.00182880000000000  \\ \hline
        TTTVT & 7 & 28 & 2667 & 6 & 2500 & 2200 & 2 & 0.25 & 0.00182880000000000  \\ \hline
        TTTVT & 38 & 12 & 2400 & 6 & 2500 & 2200 & 1 & 3.17 & 0.00384000000000000  \\ \hline
        TTTVT & 38 & 12 & 2400 & 6 & 2500 & 2200 & 2 & 3.17 & 0.00384000000000000  \\ \hline
        TTTVT & 38 & 20 & 2400 & 6 & 2500 & 2200 & 1 & 1.90 & 0.00230400000000000  \\ \hline
        TTTVT & 38 & 20 & 2400 & 6 & 2500 & 2200 & 2 & 1.90 & 0.00230400000000000  \\ \hline
        TTTVT & 38 & 24 & 2400 & 6 & 2500 & 2200 & 1 & 1.58 & 0.00192000000000000  \\ \hline
        TTTTV & 38 & 24 & 2400 & 6 & 2500 & 2200 & 2 & 1.58 & 0.00192000000000000  \\ \hline
        TTTTV & 38 & 28 & 2400 & 6 & 2500 & 2200 & 1 & 1.36 & 0.00164571428571429  \\ \hline
        TTTTV & 38 & 28 & 2400 & 6 & 2500 & 2200 & 2 & 1.36 & 0.00164571428571429  \\ \hline
        TTTTV & 38 & 12 & 2133 & 6 & 2500 & 2200 & 1 & 3.17 & 0.00341280000000000  \\ \hline
        TTTTV & 38 & 12 & 2133 & 6 & 2500 & 2200 & 2 & 3.17 & 0.00341280000000000  \\ \hline
        TTTTV & 38 & 20 & 2133 & 6 & 2500 & 2200 & 1 & 1.90 & 0.00204768000000000  \\ \hline
        TTTTV & 38 & 20 & 2133 & 6 & 2500 & 2200 & 2 & 1.90 & 0.00204768000000000  \\ \hline
        TTTTV & 38 & 24 & 2133 & 6 & 2500 & 2200 & 1 & 1.58 & 0.00170640000000000  \\ \hline
        TTTTV & 38 & 24 & 2133 & 6 & 2500 & 2200 & 2 & 1.58 & 0.00170640000000000  \\ \hline
        TTTTV & 38 & 28 & 2133 & 6 & 2500 & 2200 & 1 & 1.36 & 0.00146262857142857  \\ \hline
        TTTTV & 38 & 28 & 2133 & 6 & 2500 & 2200 & 2 & 1.36 & 0.00146262857142857  \\ \hline
    \end{tabular}}
\caption{\scriptsize The list of configurations used in $\mathbf{E}_1$. The \emph{Fold sequence} column shows whether the configuration was a part of the training (T) or validation (V) set during each cross-validation fold stated in Figure \ref{FIG:cross_validation}. For example, the sequence TTTTV indicates configuration was part of the training set during the first four folds and a part of the validation set in the fifth fold.}
\label{table:cross_validation_configs}
\end{table}

\begin{figure}
\centering
\begin{subfigure}{0.5\linewidth}
    \centering
    \includegraphics[width=\textwidth]{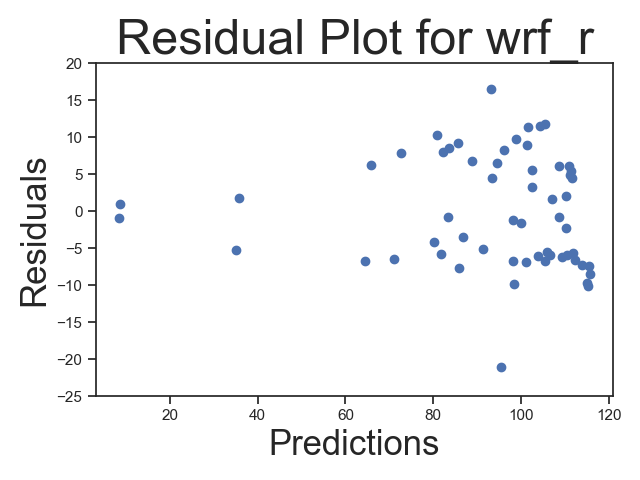}
    \caption{Residual plot: wrf\_r from $\mathbf{E}_1$, acc. $\approx88\%$}
    \label{FIG:residual_wrf_r}
\end{subfigure}%
\begin{subfigure}{0.5\linewidth}
    \centering
    \includegraphics[width=\textwidth]{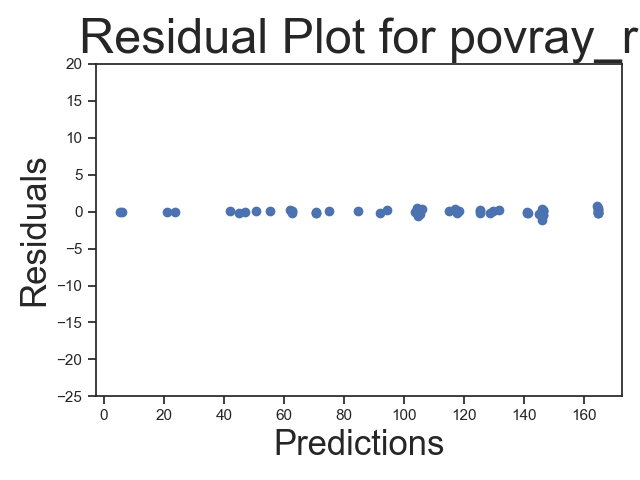}
    \caption{Residual plot: povray\_r from $\mathbf{E}_1$, acc. $>99\%$}
    \label{FIG:residual_povray_r}
\end{subfigure}
\caption{Residual plots for wrf\_r and povray\_r to demonstrate impact on model accuracy}
\label{FIG:residual_plots}
\end{figure}

While most of the above fifty eight (58) have an average cross-validated accuracy $\geq95\%$, we find that the accuracies approximately range between $80\%$ and $99\%$, thereby indicating the scope of the model performance. \added{The prediction accuracy is influenced by the accuracy of the estimates of the model parameters, which are estimated in this work by the least squares technique. The precision of the estimator is the highest when the conditions of the Gauss-Markov theorem are satisfied, thereby the estimator becomes the Best Linear Unbiased Estimator (BLUE) \cite{theil1971principles}. In a linear framework, the conditions require non-correlation and homoscedasticity of the error terms. For example, consider the least and most accurate predictions from the floating point suite of experiment $\mathbf{E}_1$. From Figure \ref{FIG:sfr_intel_results}, they belong to the wrf\_r and povray\_r tests, respectively. The residual plots of these tests are provided in Figure \ref{FIG:residual_plots}. The increasing variance of residuals with increasing values of predictions for the case of wrf\_r (Figure \ref{FIG:residual_wrf_r}) is suggestive of heteroscedasticity, thereby justifying a lower prediction accuracy ($\approx 88\%$). On the other hand, the residual behavior in the case of povray\_r (Figure \ref{FIG:residual_povray_r}) does not exhibit any tangible change in variance across the axis of predictions and is suggestive of homoscedasticity, thereby justifying a higher prediction accuracy ($>99\%$). In this manner, residual analysis is recommended to validate the assumptions pertaining to the modeling and estimation (training) procedure. It is possible to address issues related to correlation and heteroscedasticity of errors by including nonlinear features within the Amdahl's law formulation (similar to the mention in \ref{ssec:enhancements_higherorder}) and regularized estimation procedures. The reader is referred to Section 3.3.3 of \cite{james2013introduction} for details}.

In general, the model performance is not only influenced by modeling aspects such as their assumptions, structure, and training methods but also by the data \added{variation} \deleted{coverage} offered by the experiment sample. \deleted{For example, if all but one measurement have the same value for a resource, it is likely that the model would not have enough information to learn the impact of the resource on system performance.} \added{For example, consider the memory frequency resource variable. If all the data points in the sample have the same memory frequency, there is no variation in this resource variable to be able to relate it to the system performance. It follows that the model would be unable to learn the effect of the memory frequency on the system performance. In this manner, it is necessary to ensure variation in the resource variables to model their impact on the system performance.} Further, the statistical assumptions involved in least squares training need to be validated in order to build generalizable models \cite{myers1990classical}. It is also important to note that the actual execution is dynamic in nature whereas these models offer static approximations of the dynamic process to statistically relate the resource variables to the performance metric. Despite the shortcomings of analytical modeling, it results in functionally simpler models to predict system performance with reasonable accuracy as suggested by the findings of our study.
\subsection{Use cases for new designs}
\label{ssec:use_cases}
Once the models are deemed satisfactory, it is not only possible to predict performance of the benchmarks within the range of the system configurations (Table \ref{table:expt_config_ranges}) but pose the inverse problem of finding the range of resource variables for which a performance target is feasible along with cost considerations. This analysis can help in the design of new systems with desirable performance and cost characteristics.
\section{Conclusion}
\label{sec:conclusion}
In this work, the problem of predicting multicore system performance based on benchmark measurements is studied. The conventional Amdahl's law formulation is examined in the context of multicore systems. Resulting speedup equations were found to be limited to single resource enhancements with respective assumptions and do not allow for studying the simultaneous effect of multiple resource enhancements. To migitate these limitations, an extension of the Amdahl's law is proposed to incorporate the effect of multiple system resources simultaneously. A regression framework suitable for learning is derived from the extended Amdahl's law. Regression models are trained on data from experiments across multiple benchmarks and architectures. The expected prediction accuracy of these models was determined by cross-validation. Results indicate an average cross-validation accuracy of $\approx80\%-99\%$ depending on the benchmark and hardware platform considered. The predictive ability of the regression models across different benchmarks and architectures demonstrates the generalizability of the proposed Amdahl's law extension in effectively learning computer performance due to the simultaneous enhancement of multiple resources. The proposed method generalizes across various benchmarks and compute architectures. Future work should investigate the role of \deleted{experiment design}\added{optimizing experiment design in relation to estimator accuracy and efficiency}, performance reachability analysis, feature learning for rapid generation of analytical models of various product architectures and benchmarks.

\section*{Acknowledgement}
The authors thank Intel Corporation for supporting this work.

\bibliographystyle{elsarticle-num} 
{\small\bibliography{references}}





\end{document}